\documentclass[letterpaper, 10 pt, conference]{ieeeconf}
\usepackage{graphics} 
\usepackage{epsfig} 
\usepackage{multirow}
\usepackage{rotating}
\usepackage{colortbl}
\usepackage[table, dvipsnames]{xcolor}
\usepackage{makecell}
\usepackage{amssymb}
\usepackage{tabularx,booktabs}
\usepackage{mathtools}
\usepackage{graphicx}
\usepackage{caption}
\usepackage{subcaption}
\usepackage{amsmath} 
\usepackage{bbm}
\usepackage{color,soul}
\usepackage{tikz}
\usepackage{float}
\usetikzlibrary{positioning}
\usepackage{forest,array}
\usetikzlibrary{calc,shapes,shapes.arrows,arrows,trees,shadows,backgrounds,positioning}
\usepackage{multicol}
\usepackage{balance}
\usepackage{xcolor}
\usepackage[framemethod=tikz]{mdframed}
\usepackage{algpseudocode}
\usepackage{algorithm}
\usepackage{bbm}
\usepackage{dsfont}

\forestset{
  nodeStyle/.style={
    before typesetting nodes={
      edge=#1,
      for ancestors={
        edge=#1,
        #1,
      },
      #1,
    }
  },
  my edge label/.style={
    edge label={
      node [midway, fill=white, font=\scriptsize] {#1}
    }
  }
}

\newcommand{\floor}[1]{\lfloor #1 \rfloor}

\IEEEoverridecommandlockouts                              
\title{
\LARGE \bf Optimizing SLAM Evaluation Footprint Through Dynamic Range Coverage Analysis of Datasets
}

\author{
	Islam Ali$^{1}$ and Hong Zhang$^{2}$

	\thanks{$^{1}$Islam Ali is with the Department of Computing Science, University of Alberta, Edmonton, AB T6G 2R3, Canada {\tt\small iaali@ualberta.ca}}%

	\thanks{$^{2}$Hong Zhang is with the Department of Computing Science, University of Alberta, Edmonton, AB T6G 2R3, Canada {\tt\small hzhang@ualberta.ca}}%
}
\begin{document}

\maketitle
\thispagestyle{empty}
\pagestyle{empty}

\begin{abstract}
Simultaneous Localization and Mapping (SLAM) is considered an ever-evolving problem due to its usage in many applications. Evaluation of SLAM is done typically using publicly available datasets which are increasing in number and the level of difficulty. Each dataset provides a certain level of dynamic range coverage that is a key aspect of measuring the robustness and resilience of SLAM. In this paper, we provide a systematic analysis of the dynamic range coverage of datasets based on a number of characterization metrics, and our analysis shows a huge level of redundancy within and between datasets. Subsequently, we propose a dynamic programming (DP) algorithm for eliminating the redundancy in the evaluation process of SLAM by selecting a subset of sequences that matches a single or multiple dynamic range coverage objectives. It is shown that, with the help of dataset characterization and DP selection algorithm, a reduction in the evaluation effort can be achieved while maintaining the same level of coverage. We also study how the evaluation process of a real-world SLAM system can be optimized utilizing the method proposed.
\end{abstract}
\section{Introduction}
Simultaneous localization and mapping (SLAM) is the process of determining the location a mobile robot while it is building a map of the environment \cite{durrant2006simultaneous}. On the one hand, the SLAM problem was thought to be a solved problem \cite{frese2010interview} given the tasks and the controlled environments where such systems are usually deployed. On the other hand, deployment of SLAM in the wild creates an evolving problem with many open questions. For instance, the robustness and resilience of SLAM is a critical requirement for the deployment of SLAM in unknown and unstructured environments. Currently, one main open problem in SLAM is how to quantify robustness, ensure robustness, and design for robustness.  

\begin{figure}[!tp]
     \centering
         \centering
         \includegraphics[width=\columnwidth]{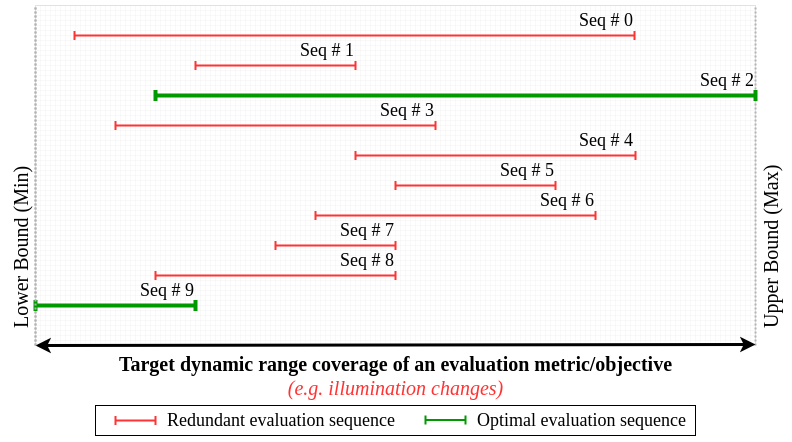}    
        \caption{An illustration of how evaluation of SLAM is usually conducted and the level of redundancy present in the process (red), and the subset of optimal evaluation sequences given a defined evaluation objective (green)}
        \label{fig:coverage_example}
\end{figure}

The robustness and resilience of SLAM cannot be determined, guaranteed, or transferred without the quantification of the design and testing conditions first, and then the validation and deployment conditions \cite{torralba2011unbiased}. The performance of SLAM is often evaluated by subjecting it to a temporal sequence of sensor measurements in the form of a pre-recorded dataset. Thus, this gives rise to the need to provide quantitative characterization of such datasets.
Usually, SLAM evaluation begins with the selection of a number of datasets where each dataset consists of a number of sequences. The selection of the datasets and their corresponding sequences are thought to capture the environment conditions and anomalies sufficiently to establish an objective evaluation of the proposed SLAM algorithm. This selection has typically been done qualitatively without an explicit consideration of the dynamic range of a certain environmental parameter. In fact, as this study proposes, there exists a huge level of redundancy and similarity among datasets and among measurement sequences in the same dataset. Thus, the analysis and identification of the coverage of dynamic range achieved by a certain experimental setup are essential in quantifying robustness of a system under test. This identification leads to the optimal selection of validation sequences to achieve the same level of coverage in terms of time and effort. With the aid of the characterization framework proposed in \cite{ali2022we}, one can identify the level of redundancy present in SLAM datasets, as shown in Figure \ref{fig:coverage_example}, and can utilize the characterization results for optimizing the evaluation process of SLAM.

In this work, we utilize the characterization results in \cite{ali2022we} to measure the coverage of dynamic range achieved by a combination of several datasets. Based on the achieved coverage, and the characterization of each measurement sequence, an optimal subset of sequences for each characterization metric is calculated. This subset ensures achieving the same level of coverage while minimizing either the number of validation sequences or the number of measurements processed. 

The rest of the paper is organized as follows. Section \ref{sec:related_work} discusses the motivation of the proposed work as well as a brief discussion of the related work. After that, Section \ref{sec:problem_def} describes the problem addressed in this work and its mathematical basis. Next, Section \ref{sec:method} describes our proposed methodology for solving the problem using a dynamic programming approach. Moreover, Section \ref{sec:results} discusses the results of the proposed method and compares its outcomes with traditional validation methods. Finally, Section \ref{sec:conclusion} presents our conclusions.
\section{Related work}
\label{sec:related_work}
With the frequent introduction of new datasets to evaluate and benchmark SLAM everyday \cite{liu2021simultaneous}, one shall ask if another dataset is needed. In fact, a more fundamental question would be which datasets to use for evaluation given a defined evaluation objective. This work provides a needed linkage between dataset characterization and single and multi-objective SLAM evaluation and benchmarking through the analysis of dynamic range coverage of datasets.

\begin{table}[!tp]
\caption{Summary of Datasets Characterization Metrics}
\label{tab:char_metrics}
\begin{tabularx}{\columnwidth}{l|X}
\toprule
\multicolumn{1}{c|}{\textbf{Group}} & \multicolumn{1}{c}{\textbf{Characterization Metrics}}                          \\ \midrule
\multicolumn{2}{c}{\textbf{General characterization metrics (G)}} \\ \midrule
Sampling and rates                  & - Samples, Total Duration, Sampling time, Sensor timestamps mismatch           \\ \midrule
\multicolumn{2}{c}{\textbf{Inertial characterization metrics (I)}} \\ \midrule
Higher-order diffs            & - Jerk, Snap, Angular acc., Angular jerk                                       \\
Sensor saturation                   & - Dynamic range covering and crossing                                          \\
Rotation-only motion                & - Acceleration magnitude, \# rotation-only samples                             \\ \midrule
\multicolumn{2}{c}{\textbf{Visual characterization metrics (V)}} \\ \midrule
Brightness                          & - Avg. brightness, Zero-mean avg. brightness derivative, Ratio of Thresholding \\
Exposure                            & - Trimmed mean, Trimmed skewness, Exposure Zone                                \\
Contrast                            & - Contrast Ratio, Weber contrast, Michelson contrast, RMS contrast             \\
Blurring                            & - Blur score, blur percentage/image, blurred images percentage                 \\
Detectable features & - Avg. \# features / sub-image, Avg. spatial dist. ratio, Abs. spatial dist. ratio             \\
Disparity                           & - Avg. \& std. dev. of disparity map (StereoBM - StereoSGBM)                   \\
Similarity                          & - DBoW2 score, distance to closet match                                        \\ \bottomrule
\end{tabularx}
\end{table}
\begin{figure*}[!tp]
     \centering
         \centering
         \includegraphics[width=\textwidth]{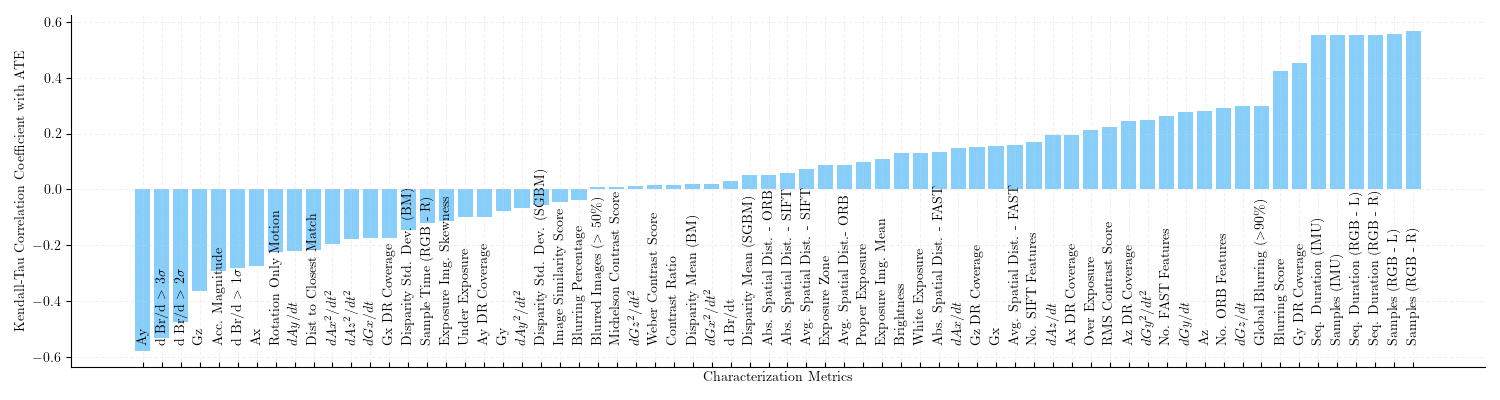}    
        \caption{Non-monotonic correlation coefficient (Kendall-Tau) between different characterization metrics of datasets and their corresponding ORB-SLAM3 absolute trajectory error(ATE)}
        \label{fig:corr_results}
\end{figure*}

Characterization of a dataset is the process of extracting descriptive metrics from the dataset systematically. In this work, we utilize the characterization metrics and results presented in \cite{ali2022we}. This framework applies several characterization metrics on each data sequence in a dataset, and produces a corresponding characterization vector. The characterization metrics are divided into three groups: general, inertial, and visual characterization metrics, which are summarized in Table \ref{tab:char_metrics}. Each characterization metric measures a single environmental property in the environment in which the dataset was recorded, and together those characterization metrics represent an abstract descriptor of the each sequence in a lower dimensional space.

The significance of the characterization metrics stems from the observed correlation between the characterization results and the corresponding \textit{Absolute Trajectory Error (ATE)}. In Figure \ref{fig:corr_results}, the non-monotonic correlation coefficient (Kendall-Tau) is measured between different characterization metrics of datasets and ORB-SLAM3 \cite{campos2021orb} ATE. We can observe medium to high correlation between those characteristics and the SLAM ATE, which suggests SLAM sensitivity to the metrics measured. Thus, coverage of such metrics must be considered in the evaluation process of SLAM.

On the other hand and similar to any single or multi-objective optimization problem \cite{deb2014multi}, the goal of SLAM algorithms is to provide acceptable localization and mapping performance w.r.t. a single or a group of objectives defined by the application in which the system will be deployed. This is manifested in several SLAM algorithms where the objective could be immunity to illumination changes \cite{park2017illumination}\cite{ranganathan2013towards}, ability to handle textureless situations \cite{dong2019novel}\cite{zhou2015structslam} or handling environment dynamics \cite{cui2019sof}\cite{zou2012coslam}\cite{brasch2018semantic}, among many others. Due to the lack of datasets characterization results of popular SLAM datasets, single-objective SLAM researchers tend to collect their own data for evaluation in order to control environmental parameters.

Not surprisingly, analysis of dataset properties is a crucial topic in a wide range of disciplines in science and engineering especially in learning problems \cite{mani2019coverage} and in pure data analysis problems \cite{asudeh2019assessing}. The analysis of dataset bias \cite{torralba2011unbiased} and dataset shift \cite{turhan2012dataset} in the aforementioned disciplines led to methods and techniques for correcting the dataset bias \cite{khosla2012undoing} particularly in the context of deep neural network models \cite{kortylewski2018can}. Although the aforementioned research is directed to other computer vision tasks, the same concepts can be adopted in SLAM research especially in learning-based SLAM \cite{macario2022comprehensive}. The concept of coverage pattern was introduced in \cite{asudeh2019assessing}, which uses characterization parameters as a feature vector. Moreover, the relationship between different patterns was modelled by edges in a directed graph. The approach proposed is suitable for situations where the characterization metrics are discrete and not continuous. In SLAM, however, the characterization of measurements is often a continuous variable. Thus, to utilize the same concepts mentioned in \cite{asudeh2019assessing}, quantization in the continuous space is required, leading to quantization errors and potential sub-optimal analysis of the coverage.

Consequently, the importance of coverage analysis naturally results from its relation to the proper design of the experimental setup used to evaluate a SLAM algorithm. This analysis was not possible due to the lack of quantitative characterization of SLAM datasets. However, with the availability of a quantitative characterization framework \cite{ali2022we}, the analysis can now be attempted. This study provides a novel approach to the coverage problem, by abstracting the characterization results of a dataset as continuous random variables to support the utilization of statistical analysis techniques and popular set manipulation algorithms benefiting from their implied optimality in selecting optimal subsets of sequences. Unlike previously mentioned studies, this work is directed toward the SLAM problem and highlighting how statistical analysis and dynamic programming can be used to optimize the SLAM evaluation process. 
\section{Problem Definition}
\label{sec:problem_def}
The existence of redundancy within and between datasets leads to inefficient design experimental setup, where evaluation objectives are not guaranteed to be addressed. This gives rise to the need to select a subset of evaluation sequences from available datasets that reaches similar coverage levels for certain quantifiable objective. For instance, we seek to answer whether the full KITTI dataset is needed to test for illumination changes or not. If not, which sequences in KITTI dataset are representing of the coverage level achieved by the full dataset. In this section, we formally describe the problem we are solving. Also, we define key parameters needed in the course of this work. 

Let $Q$ be a set of sequences of a SLAM dataset where:
\begin{equation}
Q = \{Q_i, \; i=1, ..., n\}
\end{equation}
As such, a sequence $Q_i$ is a vector of $m_i$ measurements (e.g. images/inertial measurements), i.e. $m_i = |Q_i|$, where $i$ corresponds to a sequence. 

we characterize each sequence using the method described in \cite{ali2022we}, and a corresponding characterization vector $q_i$ is generated for one of the characterization metrics mentioned in Table \ref{tab:char_metrics} so that:
\begin{equation}
q_i = \{q_{i,1}, ..., q_{i,m_i}\}
\end{equation}

$q_{i,j}$ is a feature representing the result of applying a characterization metric $f(.)$ on sample (e.g. image) $I_{i,j}$ in sequence $i$, and is given by:
\begin{equation}
q_{i,j} = f(I_{i,j})
\end{equation}

The characterization vector $q_i$ can be considered a continuous random variable that is represented by the inclusive bounded interval of minimum and maximum characterization value, as such:
\begin{equation}
\delta q_i= [\min_{j} q_{i,j}, \max_{j} q_{i,j}]
\end{equation}

Thus, the dynamic range coverage for a a given dataset of sequences $Q$ under a specific characterization metric is denoted by the interval $\Delta Q$, and is given by:
\begin{equation}
\Delta Q = (\bigcup_{i=1}^{\tilde{n}} \delta q_i)
\end{equation}
which is a bounded inclusive interval that is given by:
\begin{equation}
[\Delta Q_{min}, \Delta Q_{max}] = [\min_{i,j} q_{i,j}, \max_{i,j} q_{i,j} ]
\end{equation}
where $j$ is an index of the minimum or maximum element in the joint set $\Delta Q$.

Therefore, the outer measure \cite{rudin1976principles} of the interval $\Delta Q$ is given by: 
\begin{equation}
|\Delta Q| = |\Delta Q_{max}-\Delta Q_{min}| - \varepsilon
\end{equation}
where $\varepsilon$ is the numerical discontinuity in the $\Delta Q$.

We seek to find a subset of sequences $\tilde{Q} \subseteq Q$ of size $\tilde{n}$ such that:
\begin{equation}
\tilde{n} \leq n \;\; \& \;\; \Delta \tilde{Q} = \Delta Q
\end{equation}

To find the minimum set $\tilde{Q}$, we define two quantities to compare subsets. The first one is the cost of dynamic range coverage $C(\tilde{Q})$ defined by the number of processed measurements per unit dynamic range coverage:
\begin{equation}
C(\tilde{Q}) = ({\sum\limits_{i=1}^{\tilde{n}} m_i}) \; / \; {|\Delta \tilde{Q}|}
\end{equation}

The second is the dynamic range coverage percentage $P(\tilde{Q})$, which is the dynamic range coverage of subset $\tilde{Q}$ relative to the full set of sequences $Q$, which is given by:
\begin{equation}
P(\tilde{Q}) = \frac{|\Delta{\tilde{Q}}|}{|\Delta{Q}|} \; \%
\end{equation}

Finally, we define our problem as one of finding the subset $\tilde{Q}$ that achieves the least number of validation sequences (LS) by solving:
\begin{equation}
\begin{gathered}
\arg \min_{\tilde{n}} \;\; \tilde{n} = |\tilde{Q}| \\
s.t. \;\;\;\; \tilde{n} \leq n, \;\;\; \tilde{Q} \subseteq Q , \;\;\; \Delta \tilde{Q} = \Delta Q
\end{gathered}
\end{equation}

or the lowest possible cost (LC) of dynamic range coverage by solving:
\begin{equation}
\begin{gathered}
\arg \min_{\tilde{Q}} \;\; C(\tilde{Q}) = ({\sum\limits_{i=1}^{\tilde{n}} m_i}) \; / \; {|\Delta{\tilde{Q}}|} \\
s.t. \;\;\;\; \tilde{Q} \subseteq Q , \;\;\; \Delta \tilde{Q} = \Delta Q
\end{gathered}
\end{equation}
The two objectives aspire reducing the footprint of SLAM evaluation while maintaining the same dynamic range coverage achieved by a pool of evaluation sequences.
\begin{figure*}[!tp]
     \centering
         \includegraphics[width=\textwidth]{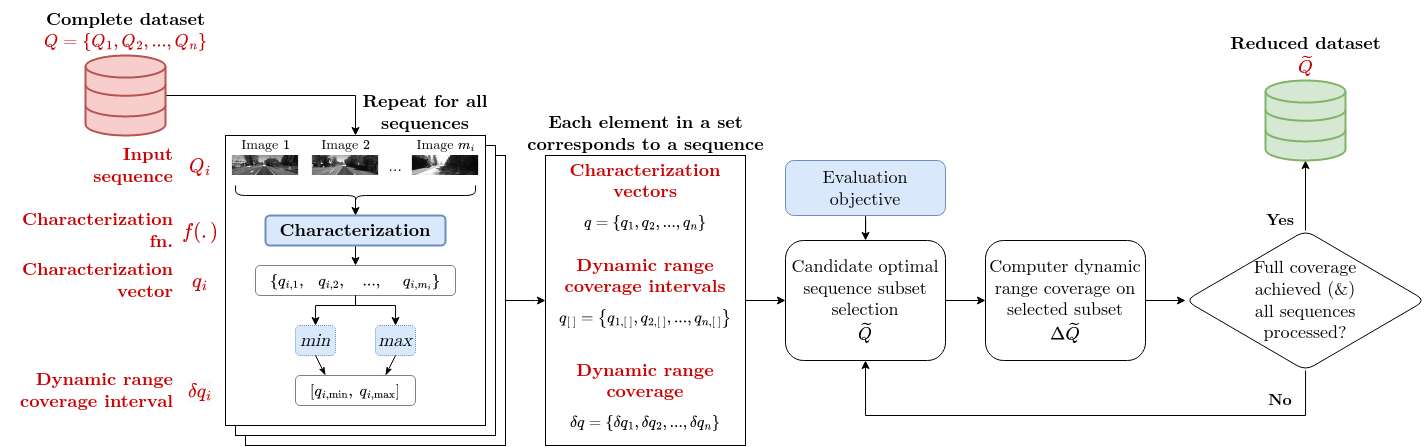}
        \caption{A block diagram of the system flow illustrating different system parameters}
        \label{fig:sys_bd}
\end{figure*}

To illustrate the relation between characterization metrics, and the aforementioned definition, we discuss the following example. Assume we want to evaluate a SLAM system against illumination changes (the objective characterization metric) on the KITTI dataset, denoted as $Q$, that consists of 22 sequences ($n=22$). After applying the illumination change characterization metric $f(.)$ on each sequence $Q_i$, we obtain a characterization vector of size $m_i$ denoted as $q_i$. We calculate the dynamic range coverage $\Delta Q$, and the cost of coverage $C(\tilde{Q})$ of processing all sequences available in KITTI. We seek to find a minimum set $\tilde{Q}$ with $\tilde{n}$ sequences. This subset $\tilde{Q}$ must achieve a dynamic range coverage for illumination changes that is equivalent to that of the full KITTI dataset by either reducing the number of processed sequences ($\tilde{n} \leq n$) or reducing the cost of dynamic range coverage ($C(\tilde{Q}) \leq C(Q)$).
\section{Methodology}
\label{sec:method}
Given a set of dataset sequences, we seek to find an optimal minimal subset of sequences that achieves a defined evaluation objective, which is achieving a dynamic range coverage that is equivalent to the full input dataset over one of characterization metrics. As shown in Figure \ref{fig:sys_bd}, the process starts by characterizing all sequences in the dataset w.r.t. the selected characterization metric using the method proposed in \cite{ali2022we}. This step produces a characterization vector and a dynamic range coverage for each data sequence. After that, this information is sent to an optimization algorithm which iteratively selects an optimal subset of sequences satisfying the evaluation criteria. At each iteration, the algorithm computes the dynamic range coverage of a candidate and updates its internal state. Once all sequences are considered, the optimal subset is reported. 

\begin{algorithm}[!tp]
	\caption{Dynamic Range Coverage Calculation}
	\label{effective_coverage_algo} 
	\textbf{Input:} $\delta q:$ List of characterization vectors
 	\\ \textbf{Output:} $\Delta Q:$ Dynamic range coverage
 	\\ \textbf{Initialization:} \\ 	
 		\hspace*{\algorithmicindent} \makebox[2cm][l]{$\rhd$ $\varepsilon$} $\gets 0$ , total discontinuity \\
 		\hspace*{\algorithmicindent} \makebox[2cm][l]{$\rhd$ $\Delta Q_{min}$} $\gets \delta q_{1,min}$ \\
 		\hspace*{\algorithmicindent} \makebox[2cm][l]{$\rhd$ $\Delta Q_{max}$} $\gets \delta q_{1,max}$	\\
 		\textbf{BEGIN:} Effective Coverage Calculation
 		\begin{algorithmic}[1] 
		\State \textbf{Sort} $\delta q$ list on min. value in characterization vector
		\For{$i$ in $range(1,n)$}
			\State $\delta$ = $[\Delta Q_{min}, \Delta Q_{max}] \cap [\delta q_{i,min}, \delta q_{i,max}]$
			\If {$\delta == 0$}
				\State $\rhd$ compute discontinuity region $\varepsilon$
				\State $t_{max} \gets max(\Delta Q_{max}, \delta q_{i,max})$
				\State $t_{min} \gets min(\Delta Q_{min}, \delta q_{i,min})$
				\State $\varepsilon\; += (t_{max} - t_{min})- |\Delta Q| - |\delta q_i|$				
			\EndIf
			\State $\Delta Q_{max} \gets max(\Delta Q_{max}, \delta q_{i,max})$
			\State $\Delta Q_{min} \gets min(\Delta Q_{min}, \delta q_{i,min})$
		\EndFor
		\State \textbf{Set} $\Delta Q = [\Delta Q_{min}, \Delta Q_{max}]$
		\State \textbf{Set} $|\Delta Q| =  \Delta Q_{max} - \Delta Q_{min} - \varepsilon$
		\State \textbf{Return:} $\Delta Q$
	\end{algorithmic} 
\end{algorithm}
Computation of the dynamic range coverage is described in Algorithm \ref{effective_coverage_algo}, which depends on integrating sequences boundaries taking into consideration numerical discontinuity regions among intervals that represent sequences in the input dataset. The algorithm yields $\Delta \tilde{Q}$ which is the dynamic range coverage of a given set of sequences $\tilde{Q}$.

On the other hand, the problem of finding the optimal subset of intervals to match a target dynamic range was historically solved using greedy-based approaches \cite{prolubnikov2015set}, due to their simplicity, resulting in an acceptable sub-optimal solution in a polynomial time.
\begin{algorithm}[!tp]
	\caption{Greedy Optimization Algorithm}
	\label{greedy_algo} 
	\textbf{Input:} $Q:$ set of sequences to optimize
	\\ \textbf{Input:} $\delta q:$ set of characterization vectors
 	\\ \textbf{Output:} $\tilde{Q}:$ optimal subset of sequences
 	\\ \textbf{Initialization:} \\ 	
 		\hspace*{\algorithmicindent} \makebox[2cm][l]{$\rhd$ $P(\tilde{Q})$} $\gets 0$ , current coverage percentage \\
 		\hspace*{\algorithmicindent} \makebox[2cm][l]{$\rhd$ $\tilde{Q}$} $\gets \{\}$ , empty set \\
 		\textbf{BEGIN:} Greedy Optimization Approach
 		\begin{algorithmic}[1] 
		\State \textbf{Sort} $\delta q$ list on min. value in characterization vector
		\State $i \gets 1$, current sequence to process
			\While{$P(\tilde{Q}) < 100 \%$}
				\If {$\delta q_{i,max} > \Delta \tilde{Q}_{max}$}
					\State \textbf{Update} $\Delta \tilde{Q}_{max} \gets \delta q_{i,max}$
				\EndIf
				\If {$\delta q_{i,min} < \Delta \tilde{Q}_{min}$}
					\State \textbf{Update} $\Delta \tilde{Q}_{min} \gets \delta q_{i,min}$
				\EndIf
				\State \textbf{Add} $Q_i$ to $\tilde{Q}$
				\State \textbf{Update} $P(\tilde{Q}) \gets (|\Delta \tilde{Q}| / |\Delta Q|) \%$
				\State \textbf{Increment} $i$ 
			\EndWhile
		\State \textbf{Return:} $\tilde{Q}$
	\end{algorithmic} 
\end{algorithm}

\textit{Greedy algorithms} is a methodology for solving optimization problems that depends on selecting the best available options at the time of decision \cite{cormen2022introduction}. The algorithm does not allow rolling back a taken decision based on observed better alternatives. Thus, it can yield a sub-optimal or even a non-optimal solution upon termination due to its reliance on achieving local-optimality as illustrated in Algorithm \ref{greedy_algo}.

As the problem definition suggests, one can abstract the problem of finding the optimal set of sub-intervals to match the range of a target interval to the famous knapsack problem \cite{salkin1975knapsack} where the optimal subset of objects are selected to fill a knapsack with defined capacity. Consequently, dynamic programming solutions can be used for the problem to achieve optimal solution with polynomial time \cite{toth1980dynamic}.

\textit{Dynamic programming (DP)} is used for solving an optimization problem by dividing it into smaller and easier sub-problems. The final optimal solution is an incremental compilation of the solution of the sub-problems \cite{cormen2022introduction}. Moreover, DP has the ability of providing optimal solutions while maintaining reasonable and linear time complexity. Tabulation-based techniques in DP depend on removing redundant calculations of sub-problems \cite{bird1980tabulation} ensuring the calculation of any sub-problem once at most as provided in Algorithm \ref{DP_algo}. The result of any sub-problem is stored in a table-like data structure and is re-used when needed. Similar to any recursive-based solution, a base case has to be defined and is used as a program entry point. In our case, the dynamic range coverage percentage $P(Q)$ is quantized into 10 regions where each represents $10\%$ coverage of the range. In addition to that, an initial empty state representing $0\%$ coverage have to be defined as a base case. Thus, the DP algorithm defines 11 states, where the first is the base state, and the last is required subset equivalent to $P(Q) = 100\%$.

In tabulation-based DP, the optimization objective is embedded in the algorithm by defining the replacement function which is responsible for replacing the current state of a table cell with another. The replacement function behaviour changes based on the optimization objective. For instance, the dynamic range coverage of the potential state solution is computed and is compared to the current state solution. The one with higher dynamic range coverage is selected. Otherwise, either the one with the least number of sequences or the least coverage cost is selected.
\begin{algorithm}[!tp]
	\caption{Dynamic Programming Algorithm}
	\label{DP_algo} 
	\textbf{Input:} $Q:$ set of sequences to optimize
	\\ \textbf{Input:} $\delta q:$ set of characterization vectors
 	\\ \textbf{Output:} $\tilde{Q}:$ optimal subset of sequences
 	\\ \textbf{Initialization:} \\ 
 		\hspace*{\algorithmicindent} \makebox[2cm][l]{$\rhd$ $T_{2..11}$} $\gets null$ \\	
 		\hspace*{\algorithmicindent} \makebox[2cm][l]{$\rhd$ $T_1$} $\gets \{\;\}$ \\
 		\textbf{BEGIN:} DP Optimization Approach
 		\begin{algorithmic}[1] 
 		\For{$i$ in $range(1,11)$}
			\For{$j$ in $n$}
				\State $L \gets \{\;\}$, temp list to current subset candidate
				\If{$T_i \; not \; null$}
					\State $L \gets T_i \cup q_j$
					\State $P(L) \gets |\Delta(L)| / |\Delta Q| \%$
					\State $loc \gets \floor{P(L)|/ 10} +1$
					\If{$loc < 11$}
						\If{$T_{loc}\; not \; null$}
							\State $T_{loc} \gets L$
						\Else
							\If{$|\Delta(T_{loc})| < |\Delta(L)|$}
								\State $T_{loc} \gets L$						
							\EndIf
							
							\If{$|\Delta(T_{loc})|=|\Delta(L)|$}
								\State \textbf{Execute:} $Replacement \; Fn.$
							\EndIf
						\EndIf
					\EndIf
				\EndIf
			\EndFor
		\EndFor
		\State \textbf{Set} $\tilde{Q} \gets T_{11}$
		\State \textbf{Return:} $\tilde{Q}$
	\end{algorithmic} 
\end{algorithm}
\section{Experimental Results and Discussion}
\label{sec:results}
In this section, we present our experimental setup, results, and discussion. This section is divided into two segments. the first segment describes how our proposed system can be used to efficiently select the validation subset of sequences given a defined evaluation criteria from a pool of validation sequences from three different datasets. On the other hand, the second segments discusses a case study where we discuss how our proposed system can help in validating and better designing evaluation processes for real world SLAM algorithms. 
\subsection{Optimal subset selection given an evaluation objective}
\begin{table*}[!tp]
\caption{Dynamic range coverage analysis of general (G), inertial (I), and visual (V) characterization metrics for optimization objective of least possible sequences (LS) and least possible cost (LC) for single dataset, all datasets, baseline, and our proposed selection method. Results are averaged over all characterization metrics in a given group.}
\centering
\label{tab:summary_res}
\begin{tabular}{lccc||ccc||ccc}
\cline{2-10}
 &
  \multicolumn{3}{c||}{\textbf{General Metrics}} &
  \multicolumn{3}{c||}{\textbf{Inertial Metrics}} &
  \multicolumn{3}{c}{\textbf{Visual Metrics}} \\ \toprule
\multicolumn{1}{l||}{\textbf{Method subset $\tilde{Q}$}}  	&
\textbf{$C(\tilde{Q})$} 				&
\textbf{$P(\tilde{Q})$}          			&
\textbf{$\tilde{n}$} 					&
\textbf{$C(\tilde{Q})$}        			&
\textbf{$P(\tilde{Q})$}          			&
\textbf{$\tilde{n}$} 					&
\textbf{$C(\tilde{Q})$}       		 	&
\textbf{$P(\tilde{Q})$}         			&
\textbf{$\tilde{n}$} \\ \toprule

\multicolumn{1}{l||}{KITTI       }                & 598433      & 17.24  & $-$       & 	$-$    & $-$  & $-$ 		 & 	3856.62  & 58.16 & $-$   			\\
\multicolumn{1}{l||}{EURO-C      }                & 6.39242e+10 & 23.96  & $-$       & 	1119.14  & 81.94  & $-$ 		 & 	1689.78  & 64.07 & $-$   			\\
\multicolumn{1}{l||}{TUM-VI      }                & 5.78004e+9  & 84.23  & $-$       & 	19954.86 & 62.37  & $N-$ 		 & 	15864.8  & 80.33 & $-$   			\\
\multicolumn{1}{l||}{KITTI \& EuroC \& TUM-VI}    & 6.32624e+9  & 100.0  & 61.0        & 	10845.5  & 100.0  & 61.0  		 & 	17969.97 & 100.0 & 61.0    			\\ \hline
\multicolumn{1}{l||}{Baseline    }                & 4.29366e+9  & 100.0  & 44.5        & 	2713.21  & 100.0  & 18.54 		 & 	14470.19 & 100.0 & 42.14   			\\ \hline
\multicolumn{1}{l||}{LS - Greedy (Ours) }         & 4.71574e+8  & 100.0  & 3.6         & 	1013.93  & 100.0  & 2.58  		 & 	2171.16  & 100.0 & 12.59   			\\
\multicolumn{1}{l||}{LS - DP (Ours)    }          & 4.7144e+8   & 100.0  & 2.6         & 	446.38   & 100.0  & 1.58  		 & 	516.74   & 100.0 & 2.78    			\\ \hline
\multicolumn{1}{l||}{LC - Greedy (Ours) }         & 3.52865e+9  & 100.0  & 12.3        & 	2063.65  & 100.0  & 11.85 		 & 	4218.88  & 100.0 & 27.51   			\\
\multicolumn{1}{l||}{LC - DP (Ours)    }          & 4.71439e+8  & 100.0  & 2.6         & 	424.53   & 100.0  & 1.58  		 & 	512.78   & 100.0 & 2.81    			\\
\bottomrule
\end{tabular}
\end{table*} 
Three datasets, KITTI \cite{geiger2012we}, EuroC-MAV \cite{burri2016euroc}, and TUM-VI \cite{schubert2018tum},  and their characterizations were used to demonstrate the performance of the proposed method. When combined, they provide a total of 61 sequences. The corresponding combined dynamic range is used as the target coverage range ($100\%$ coverage). Optimal validation subsets were generated for each characterization metric (summarized in Table \ref{tab:char_metrics}) and the associated coverage cost ($C(\tilde{Q})$) is calculated. The process was conducted twice for two different objectives: least number of sequences (LS) and least possible cost (LC).

A baseline is defined to be the process of running all available sequences in order till required coverage is achieved. In this case, the ordering of sequences is following the traditional practice of running complete datasets sequentially, and the stopping criteria is achieving dynamic range coverage equivalent to the whole pool of sequences. 

In order to show the advantage of our proposed method over traditional SLAM evaluation practice, we compare our results to using: a single dataset, the complete pool of datasets (61 sequences), and the baseline. 
\begin{table*}[!tp]
\caption{Sample optimal subset given a single metric objective divided into 3 groups: general metrics, inertial metrics, and visual metrics. }
\label{tab:sample_results}
\resizebox{\textwidth}{!}{
\begin{tabular}{@{}lcccccc@{}}
\cmidrule(l){2-7}
                                                   & \multicolumn{3}{c|}{\textbf{Least sequences (LS)}}              & \multicolumn{3}{c}{\textbf{Least cost (LC)}} \\ \cmidrule(l){2-7} 
\multicolumn{1}{c|}{\textbf{Characterization Metric}} &
  \textbf{subset $\tilde{Q}$} &
  \textbf{\# seq. $\tilde{n}$} &
  \multicolumn{1}{c|}{\textbf{Reduction \%}} &
  \textbf{subset $\tilde{Q}$} &
  \textbf{\# seq. $\tilde{n}$} &
  \textbf{Reduction \%} \\ \midrule
\multicolumn{1}{l|}{RGB samples}                   & outdoor6$^3$                 & 1 & \multicolumn{1}{c|}{98.3 \%} & outdoor6$^3$                  & 1  & 98.3 \% \\
\multicolumn{1}{l|}{IMU sample rate}               & outdoors5$^3$, corridor5$^3$ & 2 & \multicolumn{1}{c|}{96.7 \%} & outdoors5$^3$, corridor5$^3$  & 2  & 96.7 \% \\
\multicolumn{1}{l|}{IMU/RGB ts mismatch} &
  \makecell{MH\_05\_difficult$^2$, magistrale2$^3$, \\ MH\_04\_difficult$^2$} &
  3 &
  \multicolumn{1}{c|}{95.08 \%} &
  \makecell{V2\_03\_difficult$^2$, magistrale2$^3$, \\ MH\_04\_difficult$^2$} &
  3 &
  95.08 \% \\ \midrule
\multicolumn{1}{l|}{X-axis acceleration}           & MH\_01\_easy$^2$               & 1 & \multicolumn{1}{c|}{98.3 \%} & MH\_01\_easy$^2$                & 1  & 98.3 \% \\
\multicolumn{1}{l|}{X-axis angular velocity}       & room5$^3$, room4$^3$         & 2 & \multicolumn{1}{c|}{96.7 \%} & room5$^3$, room4$^3$          & 2  & 96.7 \% \\
\multicolumn{1}{l|}{Rotation only motion}          & room6$^6$                    & 1 & \multicolumn{1}{c|}{98.3 \%} & room6$^6$                     & 1  & 98.3 \% \\ \midrule
\multicolumn{1}{l|}{Avg. ORB spatial distribution} &
  corridor1$^3$, 02$^1$, corridor3$^3$, slides1$^3$ &
  4 &
  \multicolumn{1}{c|}{93.44 \%} &
  corridor1$^3$, 02$^1$, corridor3$^3$, slides2$^3$ &
  4 &
  93.44 \% \\
\multicolumn{1}{l|}{Blur percentage}               & 12$^1$, slides1$^3$          & 2 & \multicolumn{1}{c|}{96.7 \%} & 14$^1$, slides2$^3$           & 2  & 96.7 \% \\
\multicolumn{1}{l|}{Inter-image brightness change} & magistrale3$^3$              & 1 & \multicolumn{1}{c|}{98.3 \%} & magistrale3$^3$               & 1  & 98.3 \% \\ \midrule
\multicolumn{7}{l}{$^{1,2,3}$ refer to sequences from KITTI, EuroC, and TUM-VI datasets respectively.}                                                                          
\end{tabular}
}
\end{table*}
Table \ref{tab:sample_results} shows a sample of the reduced validation subsets ($\tilde{Q}$) for each of the characterization metric groups (i.e. general (G), inertial (I), and visual (V)). The same process was applied on all available characterization metrics (73 characterized metrics presented in \cite{ali2022we}) and the results were aggregated for each group of metrics, which are summarized in Table \ref{tab:summary_res}.

\begin{figure}[!tp]
     \centering
     \includegraphics[width=\columnwidth]{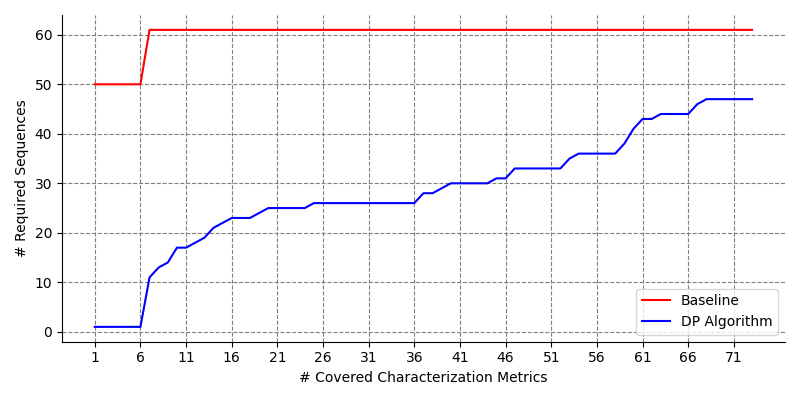}   
        \caption{The number of sequences required to achieve \textit{Joint Coverage} over a mix of characterization parameters}
        \label{fig:joint_coverage}
\end{figure}
The results show the superiority of our method in selecting the optimal subset of sequences for validation while drastically reducing the footprint of evaluation. Due to the huge redundancy in datasets, we can observe that for a single evaluation objective (e.g. addressing change in contrast or low texture environments), we only need 2 $\sim$ 3 sequences for validation compared to 61 sequences, which implies a reduction of $\approx$ 95.08 \% in evaluation efforts.

Extending this methodology to multiple objectives was conducted and presented in Figure \ref{fig:joint_coverage}. In this experiments, we aggregated the outcomes from all characterization metrics using set union to provide a preliminary indicator of joint dynamic range coverage. As shown, to cover 73 characterization metrics, only 47 sequences are needed out of 61, which implies 22.95 \% reduction in evaluation footprint, which can further be reduced by using more sophisticated techniques such as multi-objective dynamic programming.

\subsection{Case study: Illumination changes in direct SLAM}
Illumination changes in one of the challenges faced by SLAM. In \cite{park2017illumination}, this problem was addressed in direct SLAM systems, where the evaluation was conducted on a manually selected subset of public datasets. One of those datasets is TUM-RGBD \cite{sturm12iros} where five sequences were manually selected from the dataset to test for illumination changes. The question now is whether those five sequences are representing all variations in illumination and illumination changes in the dataset or not, and whether there is an optimal reduced subset that we should test against. To answer those questions, we characterized TUM-RGBD dataset using \cite{ali2022we}, and applied our DP algorithms on the characterization results for both illumination and illumination change to extract the optimal representing subset of sequences to be used for validation. 
\begin{table}[!tp]
\caption{Comparison between the manual selection and our methdology for selection with defined evaluation criterion.}
\label{tab:rgbd_results}
\resizebox{\columnwidth}{!}{
\begin{tabular}{@{}l|c|c@{}}
\toprule
\multicolumn{1}{c|}{\textbf{Selection Method}} & \textbf{Evaluation Subset $\tilde{Q}$} & \textbf{\# seq. $\tilde{n}$} \\ \midrule
Manually, used in \cite{park2017illumination}       & \makecell{fr1/desk, fr1/desk2, \\ fr1/plant, fr1/room, fr2/desk}   & 5 \\ \midrule
DP [Illumination]$^*$         & \makecell{fr2/360-kidnap, \\ fr3/nostruct-notext-near-withloop} & 2 \\ \midrule
DP [Illumination Changes]$^*$ & \makecell{fr2/360-kidnap}                                       & 1 \\ \bottomrule
\multicolumn{3}{l}{$^*$Using the method introduced in this work} \\
\multicolumn{3}{l}{\ [...] indicates the selection objective} \\
\end{tabular}
}
\end{table}
As shown in Table \ref{tab:rgbd_results}, only two sequences are needed to cover all illumination conditions, and only one sequences is needed to cover inter-image illumination changes. Additionally, we can see that the resulting subsets are completely different from what was used in \cite{park2017illumination} to represent TUM-RGBD dataset, which implies how our proposed method can help objective evaluation of SLAM given a defined criteria.

\section{Conclusions}
\label{sec:conclusion}
In this paper, the problem of measuring the dynamic range coverage of single-objective SLAM was discussed. The study started with a brief review of the topic in closely related disciplines. After that, we introduced an approach for optimizing the selection of the validation set with two different optimization objectives: minimization of the number of validation sequences, and minimization of the total cost of dynamic range coverage. The results of each optimization objective were presented and discussed. After that, the results for each characterization category were presented in detail. It was shown quantitatively that, using multiple datasets would provide more coverage than using a single dataset. However, this comes with a cost of running many validation sequences that are mostly redundant. Utilizing the DP-based approach provided a reduced mix of sequences that achieves the same coverage objectives of the whole evaluation pool of multiple datasets. The reduction achieved highlights the redundancy present in SLAM datasets, and provides a systematic approach to design SLAM experiments given defined criteria. The proposed method was used to optimize the evaluation process of a SLAM system targeting testing against illumination changes. The results show how the traditional manual selection produces non-optimal subset of sequences, while our method was able  to produce a reduced and representing subset of sequences targeting illumination change in SLAM.

The proposed method can be extend along two axes. The first is optimizing for joint coverage over a group of objectives. Initial results are presented and shows a reduction of 23 \% in footprint when 73 characterization metrics are considered. The second is using alternate sequence representation for non-uniform datasets. For the four explored datasets, the characterization metric results are uniformly distributed along their range, which enabled the utilization of interval boundary representation. However, in case of non-uniform datasets, distribution information is important to consider. The DP algorithm can be extended to accommodate both by calculating  $\Delta \tilde{Q}$, $P(\tilde{Q})$, and $C(\tilde{Q})$ over multiple ranges/objectives, while keeping the same algorithm flow. 
\bibliographystyle{IEEEtran}
\balance
\bibliography{refs}

\begin{thebibliography}{10}
\providecommand{\url}[1]{#1}
\csname url@rmstyle\endcsname
\providecommand{\newblock}{\relax}
\providecommand{\bibinfo}[2]{#2}
\providecommand\BIBentrySTDinterwordspacing{\spaceskip=0pt\relax}
\providecommand\BIBentryALTinterwordstretchfactor{4}
\providecommand\BIBentryALTinterwordspacing{\spaceskip=\fontdimen2\font plus
\BIBentryALTinterwordstretchfactor\fontdimen3\font minus
  \fontdimen4\font\relax}
\providecommand\BIBforeignlanguage[2]{{%
\expandafter\ifx\csname l@#1\endcsname\relax
\typeout{** WARNING: IEEEtran.bst: No hyphenation pattern has been}%
\typeout{** loaded for the language `#1'. Using the pattern for}%
\typeout{** the default language instead.}%
\else
\language=\csname l@#1\endcsname
\fi
#2}}

\bibitem{durrant2006simultaneous}
H.~Durrant-Whyte and T.~Bailey, ``Simultaneous localization and mapping: part
  i,'' \emph{IEEE robotics \& automation magazine}, vol.~13, no.~2, pp.
  99--110, 2006.

\bibitem{frese2010interview}
U.~Frese, ``Interview: Is slam solved?'' \emph{KI-K{\"u}nstliche Intelligenz},
  vol.~24, no.~3, pp. 255--257, 2010.

\bibitem{torralba2011unbiased}
A.~Torralba and A.~A. Efros, ``Unbiased look at dataset bias,'' in \emph{CVPR
  2011}.\hskip 1em plus 0.5em minus 0.4em\relax IEEE, 2011, pp. 1521--1528.

\bibitem{ali2022we}
I.~Ali and H.~Zhang, ``Are we ready for robust and resilient slam? a framework
  for quantitative characterization of slam datasets,'' in \emph{2022 IEEE/RSJ
  International Conference on Intelligent Robots and Systems (IROS)}, 2022, pp.
  2810--2816.

\bibitem{liu2021simultaneous}
Y.~Liu, Y.~Fu, F.~Chen, B.~Goossens, W.~Tao, and H.~Zhao, ``Simultaneous
  localization and mapping related datasets: A comprehensive survey,''
  \emph{arXiv preprint arXiv:2102.04036}, 2021.

\bibitem{campos2021orb}
C.~Campos, R.~Elvira, J.~J.~G. Rodr{\'\i}guez, J.~M. Montiel, and J.~D.
  Tard{\'o}s, ``Orb-slam3: An accurate open-source library for visual,
  visual--inertial, and multimap slam,'' \emph{IEEE Transactions on Robotics},
  vol.~37, no.~6, pp. 1874--1890, 2021.

\bibitem{deb2014multi}
K.~Deb, ``Multi-objective optimization,'' in \emph{Search methodologies}.\hskip
  1em plus 0.5em minus 0.4em\relax Springer, 2014, pp. 403--449.

\bibitem{park2017illumination}
S.~Park, T.~Sch{\"o}ps, and M.~Pollefeys, ``Illumination change robustness in
  direct visual slam,'' in \emph{2017 IEEE international conference on robotics
  and automation (ICRA)}.\hskip 1em plus 0.5em minus 0.4em\relax IEEE, 2017,
  pp. 4523--4530.

\bibitem{ranganathan2013towards}
A.~Ranganathan, S.~Matsumoto, and D.~Ilstrup, ``Towards illumination invariance
  for visual localization,'' in \emph{2013 IEEE International Conference on
  Robotics and Automation}.\hskip 1em plus 0.5em minus 0.4em\relax IEEE, 2013,
  pp. 3791--3798.

\bibitem{dong2019novel}
Y.~Dong, S.~Wang, J.~Yue, C.~Chen, S.~He, H.~Wang, and B.~He, ``A novel
  texture-less object oriented visual slam system,'' \emph{IEEE Transactions on
  Intelligent Transportation Systems}, vol.~22, no.~1, pp. 36--49, 2019.

\bibitem{zhou2015structslam}
H.~Zhou, D.~Zou, L.~Pei, R.~Ying, P.~Liu, and W.~Yu, ``Structslam: Visual slam
  with building structure lines,'' \emph{IEEE Transactions on Vehicular
  Technology}, vol.~64, no.~4, pp. 1364--1375, 2015.

\bibitem{cui2019sof}
L.~Cui and C.~Ma, ``Sof-slam: A semantic visual slam for dynamic
  environments,'' \emph{IEEE access}, vol.~7, pp. 166\,528--166\,539, 2019.

\bibitem{zou2012coslam}
D.~Zou and P.~Tan, ``Coslam: Collaborative visual slam in dynamic
  environments,'' \emph{IEEE transactions on pattern analysis and machine
  intelligence}, vol.~35, no.~2, pp. 354--366, 2012.

\bibitem{brasch2018semantic}
N.~Brasch, A.~Bozic, J.~Lallemand, and F.~Tombari, ``Semantic monocular slam
  for highly dynamic environments,'' in \emph{2018 IEEE/RSJ International
  Conference on Intelligent Robots and Systems (IROS)}.\hskip 1em plus 0.5em
  minus 0.4em\relax IEEE, 2018, pp. 393--400.

\bibitem{mani2019coverage}
S.~Mani, A.~Sankaran, S.~Tamilselvam, and A.~Sethi, ``Coverage testing of deep
  learning models using dataset characterization,'' \emph{arXiv preprint
  arXiv:1911.07309}, 2019.

\bibitem{asudeh2019assessing}
A.~Asudeh, Z.~Jin, and H.~Jagadish, ``Assessing and remedying coverage for a
  given dataset,'' in \emph{2019 IEEE 35th International Conference on Data
  Engineering (ICDE)}.\hskip 1em plus 0.5em minus 0.4em\relax IEEE, 2019, pp.
  554--565.

\bibitem{turhan2012dataset}
B.~Turhan, ``On the dataset shift problem in software engineering prediction
  models,'' \emph{Empirical Software Engineering}, vol.~17, no.~1, pp. 62--74,
  2012.

\bibitem{khosla2012undoing}
A.~Khosla, T.~Zhou, T.~Malisiewicz, A.~A. Efros, and A.~Torralba, ``Undoing the
  damage of dataset bias,'' in \emph{European Conference on Computer
  Vision}.\hskip 1em plus 0.5em minus 0.4em\relax Springer, 2012, pp. 158--171.

\bibitem{kortylewski2018can}
A.~Kortylewski, B.~Egger, A.~Morel-Forster, A.~Schneider, T.~Gerig, C.~Blumer,
  C.~Reyneke, and T.~Vetter, ``Can synthetic faces undo the damage of dataset
  bias to face recognition and facial landmark detection?'' \emph{arXiv
  preprint arXiv:1811.08565}, 2018.

\bibitem{macario2022comprehensive}
A.~Macario~Barros, M.~Michel, Y.~Moline, G.~Corre, and F.~Carrel, ``A
  comprehensive survey of visual slam algorithms,'' \emph{Robotics}, vol.~11,
  no.~1, p.~24, 2022.

\bibitem{rudin1976principles}
W.~Rudin \emph{et~al.}, \emph{Principles of mathematical analysis}.\hskip 1em
  plus 0.5em minus 0.4em\relax McGraw-hill New York, 1976, vol.~3.

\bibitem{prolubnikov2015set}
A.~Prolubnikov, ``The set-cover problem with interval weights and the greedy
  algorithm for its solution,'' \emph{Comput. Technologies}, vol.~20, no.~6,
  pp. 70--84, 2015.

\bibitem{cormen2022introduction}
T.~H. Cormen, C.~E. Leiserson, R.~L. Rivest, and C.~Stein, \emph{Introduction
  to algorithms}.\hskip 1em plus 0.5em minus 0.4em\relax MIT press, 2022.

\bibitem{salkin1975knapsack}
H.~M. Salkin and C.~A. De~Kluyver, ``The knapsack problem: a survey,''
  \emph{Naval Research Logistics Quarterly}, vol.~22, no.~1, pp. 127--144,
  1975.

\bibitem{toth1980dynamic}
P.~Toth, ``Dynamic programming algorithms for the zero-one knapsack problem,''
  \emph{Computing}, vol.~25, no.~1, pp. 29--45, 1980.

\bibitem{bird1980tabulation}
R.~S. Bird, ``Tabulation techniques for recursive programs,'' \emph{ACM
  Computing Surveys (CSUR)}, vol.~12, no.~4, pp. 403--417, 1980.

\bibitem{geiger2012we}
A.~Geiger, P.~Lenz, and R.~Urtasun, ``Are we ready for autonomous driving? the
  kitti vision benchmark suite,'' in \emph{2012 IEEE conference on computer
  vision and pattern recognition}.\hskip 1em plus 0.5em minus 0.4em\relax IEEE,
  2012, pp. 3354--3361.

\bibitem{burri2016euroc}
M.~Burri, J.~Nikolic, P.~Gohl, T.~Schneider, J.~Rehder, S.~Omari, M.~W.
  Achtelik, and R.~Siegwart, ``The euroc micro aerial vehicle datasets,''
  \emph{The International Journal of Robotics Research}, vol.~35, no.~10, pp.
  1157--1163, 2016.

\bibitem{schubert2018tum}
D.~Schubert, T.~Goll, N.~Demmel, V.~Usenko, J.~St{\"u}ckler, and D.~Cremers,
  ``The tum vi benchmark for evaluating visual-inertial odometry,'' in
  \emph{2018 IEEE/RSJ International Conference on Intelligent Robots and
  Systems (IROS)}.\hskip 1em plus 0.5em minus 0.4em\relax IEEE, 2018, pp.
  1680--1687.

\bibitem{sturm12iros}
J.~Sturm, N.~Engelhard, F.~Endres, W.~Burgard, and D.~Cremers, ``A benchmark
  for the evaluation of rgb-d slam systems,'' in \emph{Proc. of the
  International Conference on Intelligent Robot Systems (IROS)}, Oct. 2012.

\end{thebibliography}

\end{document}